\documentclass[a4paper, 10pt, conference]{ieeeconf}      

\IEEEoverridecommandlockouts                              

\usepackage{graphics} 
\usepackage{epsfig} 
\usepackage{mathptmx} 
\usepackage{times} 
\usepackage{amsmath} 
\usepackage{amssymb}  
\usepackage{tikz}
\usepackage[utf8]{inputenc}
\usepackage{pgfplots} 
\usepackage{pgfgantt}
\usepackage{pdflscape}
\pgfplotsset{compat=newest} 
\pgfplotsset{plot coordinates/math parser=false}

\usepackage{cite}
\usepackage{amsmath,amssymb,amsfonts}
\usepackage{algorithm, algorithmic}
\usepackage{graphicx}
\usepackage{textcomp}
\usepackage{xcolor}
\usepackage{svg}
\usepackage{array}
\usepackage{booktabs}
\usepackage{optidef}
\usepackage{afterpage}
\usepackage{bm}
\usepackage{gensymb}
\usepackage{colortbl}
\usepackage{color}
\usepackage{tabularx}
\usepackage{hyperref}

\usepackage{pgfplots}
\pgfplotsset{compat=newest}
\usepackage{tikzscale}

\definecolor{lred}{rgb}{1,0.647,.647}
\definecolor{orange}{rgb}{.93, .49, .19}
\definecolor{yellow}{rgb}{1,1,0.647}
\definecolor{green}{rgb}{0.647,1,0.647}
\definecolor{lblue}{rgb}{.27, .45, .76}

\usepackage{makecell}

\title{\LARGE \bf
Passive Shape Locking for Multi-Bend Growing Inflated Beam Robots}

\author{Rianna Jitosho*, Sofia Sim\'{o}n-Trench*, Allison M. Okamura, and Brian H. Do
\thanks{*These authors contributed equally to this work.}%
\thanks{This work was supported in part by National Science Foundation grant 2024247, a National Science Foundation Graduate Research Fellowship, the U.S. Department of Energy, National Nuclear Security Administration, Office of Defense Nuclear Nonproliferation Research and Development (DNN R\&D) under subcontract from Lawrence Berkeley National Laboratory; and the United States Federal Bureau of Investigation contract 15F06721C0002306.}
\thanks{The authors are with the Dept. of Mechanical Engineering, Stanford University, Stanford, CA 94305, USA. Email: \{rjitosho, sofiast, aokamura, brianhdo\}@stanford.edu}%
}

\begin{document}
\maketitle
\thispagestyle{empty}
\pagestyle{empty}

\begin{abstract} Shape change enables new capabilities for robots. One class of robots capable of dramatic shape change is soft growing ``vine" robots. These robots usually feature global actuation methods for bending that limit them to simple, constant-curvature shapes. Achieving more complex ``multi-bend" configurations has also been explored but requires choosing the desired configuration ahead of time, exploiting contact with the environment to maintain previous bends, or using pneumatic actuation for shape locking. In this paper, we present a novel design that enables passive, on-demand shape locking. Our design leverages a passive tip mount to apply hook-and-loop fasteners that hold bends without any pneumatic or electrical input. We characterize the robot’s kinematics and ability to hold locked bends. We also experimentally evaluate the effect of hook-and-loop fasteners on beam and joint stiffness. Finally, we demonstrate our proof-of-concept prototype in 2D. Our passive shape locking design is a step towards easily reconfigurable robots that are lightweight, low-cost, and low-power.
\end{abstract}

\section{Introduction}

Robots traditionally feature a fixed morphology incapable of changing after design. However, in many real-world applications it is advantageous to have robots that can change their shape to adapt to tasks rather than being immutable. 

Inflatable robots inherently offer some reconfigurability, from a compact, stowed state to a deployed state. In this work, we focus on ``vine" robots, a class of growing inflated beam robots previously developed for exploration~\cite{der2021roboa, Luong2019, Naclerio2018,CoadRAM2020, GreerSoRo2019} and manipulation~\cite{BlumenscheinROBOSOFT2018, StroppaICRA2020,  JeongIROS2020}. They are capable of significant length change by ``growing" via tip eversion. Vine robots are also capable of dramatic shape change~\cite{BlumenscheinROBOSOFT2018}.

Many implementations of vine robots feature global bending actuators such as cables or pneumatic muscles routed along the length of the robot~\cite{GreerSoRo2019, CoadRAM2020, GreerICRA2017, BlumenscheinROBOSOFT2018}. These actuators shorten one side of the vine robot, resulting in bending along the length of the entire robot. However, they are only able to produce a single, constant-curvature bend~\cite{GreerSoRo2019, CoadRAM2020, GreerICRA2017, BlumenscheinROBOSOFT2018}. The ability to form multiple bends along the length of the vine body increases its dexterity. In this work, we present a design that achieves this by pairing global bending actuators with hook-and-loop fasteners (commonly known as Velcro). By using hook-and-loop fasteners to enforce strain limits, we can control which regions of the vine robot can be shortened by the bending actuators. While there are other existing methods for achieving multiple bends that we discuss in Sec.~\ref{sec:prior_work} \cite{ AghareseICRA2018, SladeIROS2017,Selvaggio2020, GreerIJRR2020,HawkesScienceRobotics2017,Wang2020, DoICRA2020}, our design 1) enables multi-bending on demand, 2) achieves passive shape locking without relying on environmental interactions, and 3) is easy to fabricate and reset during use. To the best of our knowledge, our design is the first to combine all of these features. 

We see our work as a step towards completely reconfigurable structures. One potential area where our work could be applied is reconfigurable inflated deployable space structures~\cite{Viquerat2015, Peypoudat2005}. Another area is construction, where rapid inflated structures could be fabricated on-site without the use of traditional fabrication materials~\cite{VanDessel2003}.

\begin{figure}[t]
    \centering
    \includegraphics[width=\linewidth]{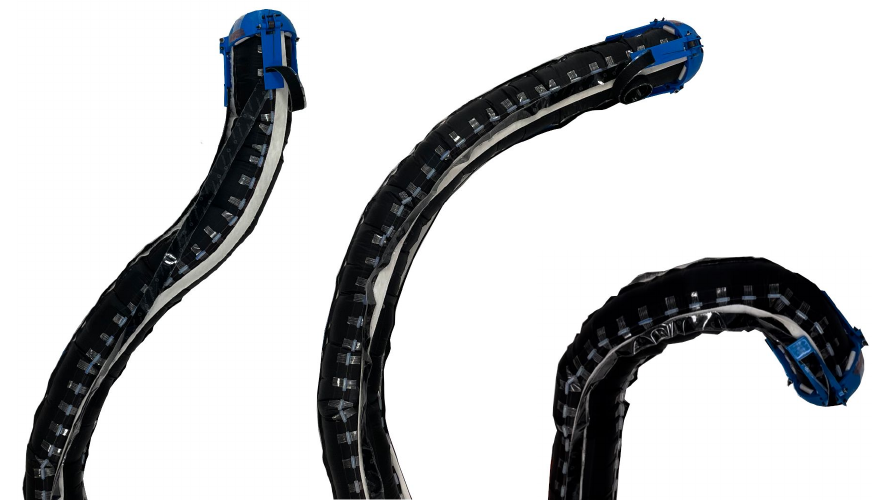}
    \vspace{-5mm}
    \caption{Photos showing various final robot configurations possible with our proposed shape locking design.}
    \label{fig:spash_photo}
    \vspace{-3mm}
\end{figure}

\begin{figure}[t]
    \centering
    \vspace{1mm}
     \includegraphics[width=3.3in]{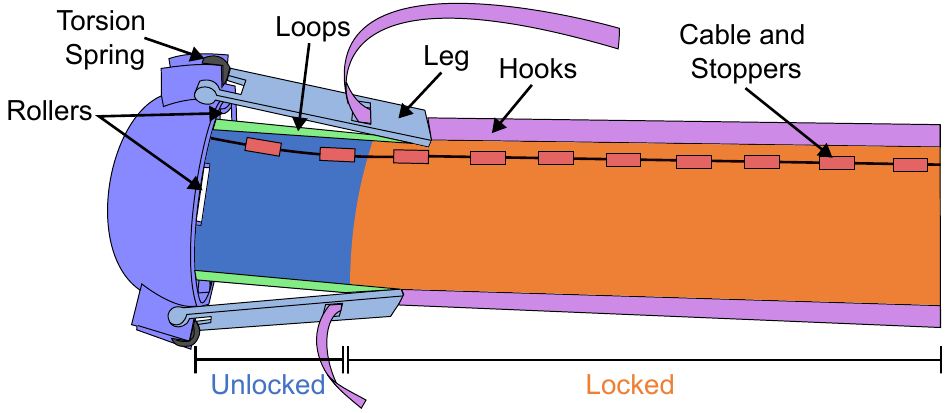}
     \vspace{-3mm}
    \caption{Overall system design. Pulling on the cable would cause bending in the blue unlocked region. Growing the vine via tip eversion would push the tip mount forward and apply hooks onto the currently exposed loops.}
    \label{fig:overall_design}
    \vspace{-3mm}
\end{figure}

The contributions of our work are as follows:
\begin{enumerate}
    \item A design for multi-bend growing robots that allows choosing a deployed shape on demand, achieves passive shape locking without relying on external contact forces, and is easy to fabricate and reset.
    \item Models and experimental characterizations that provide guidelines for implementing our design.
    \item Demonstrations on a physical prototype validating the performance of our shape locking design integrated with a vine robot.
\end{enumerate}
Fig.~\ref{fig:spash_photo} shows example deployments of our prototype. Fig.~\ref{fig:overall_design} shows an overview of our system.

\section{Prior Work on Multi-Bend Vine Robots}
\label{sec:prior_work}

Previous work on achieving multiple bends for vine robots can be categorized into three general strategies: preformed bending, contact-based bending, and shape locking. Here we provide key points of comparison relevant to our work; Table~\ref{tab:designs} summarizes these points. Additional details about these methods can be found in Blumenschein et al.~\cite{BlumenscheinFrontiers2020}.

\begin{table}[b]
    \centering
    \small
    \caption{Comparison of Soft Growing Multi-Bend Robots}
    \vspace{-3mm}
    \begin{tabular}{|>{\centering\arraybackslash}m{12mm}|>{\centering\arraybackslash}m{22mm}|>{\centering\arraybackslash}m{10mm}|>{\centering\arraybackslash}m{13.5mm}|>{\centering\arraybackslash}m{7mm}|}
    \hline
     & \textbf{Preformed  \cite{HawkesScienceRobotics2017, AghareseICRA2018, SladeIROS2017}~/ Contact-based \cite{Selvaggio2020, GreerIJRR2020}}  & \textbf{Latches \cite{HawkesScienceRobotics2017}} & \textbf{Active shape locking \cite{Wang2020, DoICRA2020}} & \textbf{Ours} \\
    \hline
    Configure on demand? & \cellcolor{lred} No & \cellcolor{green} Yes & \cellcolor{green} Yes & \cellcolor{green} Yes\\
    \hline
    Passive locking? & \cellcolor{green} Yes & \cellcolor{green} Yes & \cellcolor{lred} No & \cellcolor{green} Yes\\
    \hline
    Ease of fabrication? & \cellcolor{green} Easy & \cellcolor{lred} Difficult & \cellcolor{yellow} Moderate & \cellcolor{green} Easy\\
    \hline
    \end{tabular}
    \label{tab:designs}
\end{table}

Preformed bending involves choosing the desired configuration ahead of time. Typically, this is achieved by pinching material at desired bend locations along the vine body such that the inflated shape bends to match a desired configuration. One method for holding these pinches in place is to tape over the pinch~\cite{HawkesScienceRobotics2017}. Another option for vine robots made from thermoplastics is to heat, reshape, and cool the body material such that the deployed and desired configurations match~\cite{AghareseICRA2018, SladeIROS2017}. Vine robots that utilize preformed bending are typically easy to fabricate and allow for passive multi-bending, but require the user to commit to a specific deployed configuration ahead of time.

Contact-based bending leverages contact with the environment to maintain previous bends. Contact-based vine robots use simple, constant-curvature bending actuation~\cite{Selvaggio2020},
or carefully chosen preformed bends~\cite{GreerIJRR2020}, and then utilize contact forces to achieve multi-bend configurations. Both examples are efficient in that they use the environment rather than additional actuation to achieve multi-bending. This leads to simpler, lighter, and cheaper systems. However, their ability to create multiple bends is dependent on being deployed in an environment that provides the necessary contact forces. In addition, the work in~\cite{GreerIJRR2020} required knowledge of the environment and desired configuration a priori.

Shape locking involves maintaining multi-bends without requiring external elements. Passive shape locking has previously been achieved by a series of latches around and along the vine robot body~\cite{HawkesScienceRobotics2017}. These latches held pinches of material, and when opened due to pneumatic pressure, would cause asymmetric lengthening at the tip and thus steer the tip of the vine robot. While effective and simple during deployment, these vine robots were time consuming to manufacture and reset for subsequent use, making them impractical. There has also been work on active shape locking. These methods are able to ``lock" or ``unlock" regions of the robot body such that global bending actuators only cause bending in the unlocked regions. Wang et al. used pressurized chambers that can grow along the sides of a vine robot~\cite{Wang2020}. These chambers grow independently from the main vine body such that the vine robot is shape-locked from its base to the tip of the chambers. Do et al. fabricated a vine with segments that stiffen via layer jamming~\cite{DoICRA2020}. In this method, discrete bending occurs at the unstiffened sections of the vine. Both of these active shape locking designs enable environment-independent multi-bending but require additional actuation to maintain their bends.

Our proposed design addresses the drawbacks of each prior strategy for multi-bending. First, our design allows the robot configuration to be chosen at the time of deployment. Second, our design achieves multi-bending without relying on contact from the environment. Third, our design locks its shape passively. Finally, our design is easy to fabricate and simple to reset between deployments. 

\section{Design and Implementation}
\label{sec:design}

Here, we describe the details of the shape locking concept, passive tip mount, and fabricated proof-of-concept prototype.

\begin{figure}[t]
    \centering
    \includegraphics[width=2.6in]{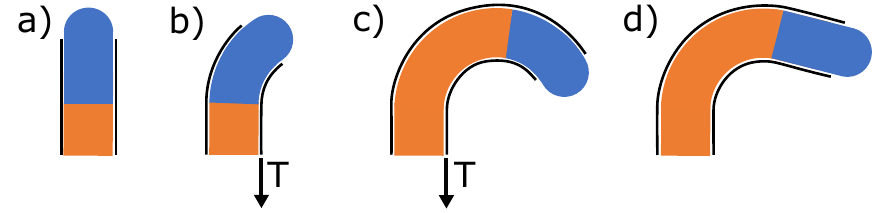}
    \vspace{-3mm}
    \caption{Conceptual illustration showing how we achieve passive shape locking. Sections in blue are unlocked, sections in orange are locked, and our tip mount is designed such that the most distal section remains unlocked. Cables are shown in black, and applying cable tension T causes bending of the unlocked portion of the vine. Growing the vine (via tip eversion) while applying cable tension will lock the bend.}
    \label{fig:locking_concept}
    \vspace{-3mm}
\end{figure}

\subsection{Utilizing Hook-and-Loop Fasteners for Holding Bends}
Our proposed design utilizes hook-and-loop fasteners, and in this paper, we refer to the constituent parts of these fasteners as ``hooks" and ``loops". Our design applies hooks onto loops on the vine body at a set distance from the tip, leaving the entire length locked except for the most distal region. Leaving the most distal region unlocked enables our bending actuators (cables) to bend only the unlocked region, even though the cables route along the full length of the vine. Fig.~\ref{fig:locking_concept} illustrates this. In a), the vine grows straight. In b), cable tension $T$ is applied to one side, which causes bending in the unlocked section of the vine (blue). In c), the vine continues growing with the tension maintained, which forms and locks a bend. In d), the cable tension is released and the proximal portion of the previous bend (orange) remains locked. This process can be repeated to form additional bends, and previously locked bends will remain in place. 

We leverage three key characteristics of hook-and-loop fasteners to achieve this. First, these fasteners have indeterminate match-up between the hooks and loops. Thus, any part of the hooks can engage with any part of the loops, enabling the freedom to pinch material anywhere along the vine body for forming bends. Second, these fasteners are easily engaged but difficult to disengage, enabling the robot to apply fasteners easily and hold bends passively. Third, the loops are flexible enough such that when attached to the vine body they do not hinder robot eversion.

\subsection{Tip Mount for Passive Fastener Application}
We designed an external tip mount that is able to passively apply the hook-and-loop fasteners onto the vine, shown in Fig.~\ref{fig:overall_design}. Our tip mount is pushed forward by the eversion of the vine. There are three legs corresponding to three lines of loops on the vine. Each leg has a slot to guide the  hooks onto the loops. The hooks are then pressed down with the far edge of the leg. A torsion spring pushes the legs inward to maintain contact with the vine. The legs apply the hooks to the loops at a fixed distance away from the tip, which allows the robot to have its most distal region unlocked and able to bend. The length of the legs determines the length of the unlocked region. The rollers on the tip mount reduce the friction between the outside vine material and the tip mount, allowing vine material to evert easily. 

\subsection{System Implementation}
We fabricated a complete vine robot system based on the components described in the previous sections. The vine body is 40-denier thermoplastic polyurethane (TPU)-coated ripstop nylon that is sealed using an ultrasonic welder. The diameter of the vine is 10.8~cm, which provides sufficient surface area to attach the fasteners and cables. The length is 2.3~m to ensure there is enough length to make more than one bend. The loops (McMaster 9652K167) are attached to the outside of the vine with double sided tape. For bending actuation, there are plastic stoppers and cables (spectra fiber braided fishing line) routed through them. Stoppers are used on the outside of the vine to hold the cable to the body. The stoppers are cut from 5~mm outer diameter Teflon tubing and taped onto the vine. Having more and shorter stoppers creates a smoother curve than fewer and longer stoppers, but this increases the fabrication time. In our implementation, the stoppers are 19~mm long and are spaced 19~mm apart, allowing for a maximum contraction ratio of 0.5.

The tip mount is 3D printed with polylactic acid (PLA) filament. The leg length was chosen based on empirical tests. A minimum length is needed to provide an unlocked region, but longer legs become cumbersome. There is a 270$\degree$ torsion spring to ensure that the legs have sufficient range of motion to maintain contact with the vine body for all possible bend angles. We chose a spring constant that would ensure the legs applied pressure to the vine without causing noticeable deformation. The depth of the tip mount was chosen empirically such that it was deep enough to capture the tip, but short enough so as not to impede vine bending. The unattached section of the hooks is coiled (Fig.~\ref{fig:demo1}), and they are pulled by the tip mount during deployment. 



\section{Modeling}
In this section, we present models that describe capabilities for any general implementation of our proposed shape locking design. First, we describe the relationship between relevant design parameters and the minimum possible radius of curvature for bends created during deployment. These kinematic relationships allow 1)~the designer to set parameters according to the curvature of bends they will need to create and 2)~the human operator to understand what is required to achieve a desired configuration. Second, we describe a relationship that enables the designer to understand the trade-offs between higher bend curvature and higher beam stiffness when implementing our design.

\subsection{Kinematics}
\label{sec:kinematics}

\begin{figure}[b]
    \centering
    \vspace{-1mm}
    \includegraphics[width=1in]{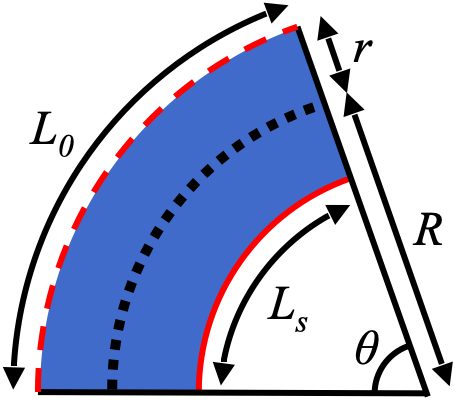}
    \vspace{-2mm}
    \caption{Relevant geometric values for solving robot kinematics. For an inflated beam with initial length $L_0$ and radius $r$, pulling a cable until all stoppers (in red) are in contact yields the contracted length $L_s$. The resulting bend has angle $\theta$ and radius of curvature $R$.}
    \vspace{-0mm}
    \label{fig:kinematics}
\end{figure}

To describe the kinematics of our multi-bending vine, we first consider the bending of an inflated beam with cables and stoppers. The relevant geometric parameters are illustrated in Fig.~\ref{fig:kinematics}. To achieve the maximum bend angle $\theta$ (and minimum radius of curvature $R$), we pull on one cable until all its stoppers are in contact. If each stopper has length $l_s$ and the gap between adjacent stoppers is length $l_g$, then we define the contraction ratio $a$ as:
\begin{equation}
    \label{eq:contraction_ratio}
    a \coloneqq \frac{L_0-L_S}{L_0} = \frac{l_g}{l_s+l_g}
\end{equation}
\noindent where $L_0$ is the original length of the inflated beam and $L_S$ is its shortened length due to pulling on the cable. Using the relationship for arc lengths and the associated subtended angles, we can form the following two relationships:
\begin{align}
    L_0 = (R+r)\theta \\
    L_S = (R-r)\theta
\end{align}
where $R$ is the bend's radius of curvature, $r$ is the radius of the inflated beam, and $\theta$ is the bend angle. By combining these equations with Eq.~\ref{eq:contraction_ratio}, we have the following model for the bend's radius of curvature:
\begin{equation}
    \label{eq:R}
    R = r \frac{2-a}{a}.
\end{equation}
Intuitively, with higher contraction ratios we can make tighter bends, and we can tune $a$ according to the highest curvature bend we expect the system to make during use. However, we are limited in that $l_s, l_g \geq 0$ for fabrication, which means $0 \leq a \leq 1$ (Eq.~\ref{eq:contraction_ratio}). This results in $R \geq 2r$ (Eq.~\ref{eq:R}).

To apply this relationship to the full multi-bending vine, consider what happens during deployment. As a simple strategy for achieving a desired configuration, the vine can either be grown with zero tension in the cables or with cable tension $T$ that results in all the stoppers in the unlocked region being in contact. Thus, the overall shape can be composed of a series of bends, each with radius of curvature $R$, and straight line segments. To achieve some desired bend angle $\theta$, we would grow the vine by length $L = (R+r) \theta$ while maintaining the cable tension $T$.

\subsection{Bend-Holding Capability}
\label{subec:bend_holding}

\begin{figure}[t]
    \centering
    \includegraphics[width=3.1in]{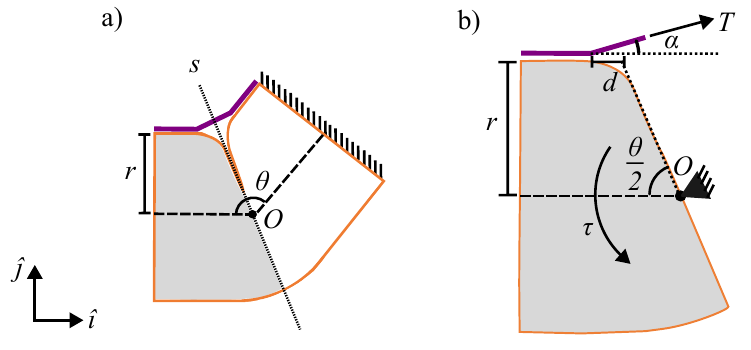}
    \vspace{-3mm}
    \caption{Relevant parameters for an inflated beam (orange) with a bend held by a hook-and-loop fastener (purple). a) Bent inflated beam fixed at one end. b) Free body diagram for half of an inflated beam. There exists a resistance torque $\tau$ due to bending the beam and a tension $T$ from the hooks.}
    \vspace{-4mm}
    \label{fig:bending_fbd}
\end{figure}

An inflated beam increases in stiffness when its internal pressure increases. However, inflated beams also generate a resistance torque when bent which acts to straighten out the beam. This torque scales with internal pressure. As a result, there exists a design trade-off wherein we would like to operate our vine at higher internal pressures for added stiffness, but we still require the hook-and-loop fasteners to overcome the resistance torque to lock bends.

We consider an inflated beam with a bend locked by hook-and-loop fasteners (Fig.~\ref{fig:bending_fbd}) since this is comparable to a short section of our robot. The bend angle is $\theta$, beam radius is $r$, and center of rotation is $O$. We assume one end of the beam is fixed and find a model that describes the pressure required to cause fastener separation. For this, we consider one half of the beam as shown in Fig.~\ref{fig:bending_fbd}, and treat $O$ as a pin joint. There exists a tension $T$ from the hook-and-loop fastener as well as a resistance torque $\tau$ that tries to straighten out the bent beam.
Before fastener separation, we have moment balance about $O$:
\begin{equation}
\label{eq:moment_balance}
    \tau - rT\cos{\alpha} - \left(\frac{r}{\tan\frac{\theta}{2}}+d\right)T\sin{\alpha} = 0
\end{equation}
\noindent where $d$ is the distance shown in Fig.~\ref{fig:bending_fbd} and $\alpha$ is the angle of $T$ with respect to horizontal. Assuming symmetry about the line $s$ in Fig.~\ref{fig:bending_fbd}a, $\alpha = \frac{\pi-\theta}{2}$. Nesler et al. provides a closed-form expression for the resistance torque $\tau$ that results from bending an inflated beam:
\begin{equation}
\label{eq:pressure}
    \tau = P \frac{dV(\theta)}{d\theta}= -\pi r^3 P \left(\tan^2\frac{\theta}{2} + 1\right)
\end{equation}
\noindent where $P$ is the gauge pressure, $V$ is the volume, $r$ is the radius, and $\theta$ is the angular deflection of the beam~\cite{Nesler2018}. 

We evaluate the maximum possible tension prior to fastener separation by using a strength criterion similar to that presented by Salama et al.~\cite{Salama2002}:
\vspace{-.1mm}
\begin{equation}
\label{eq:strength_criterion}
    \left(\frac{\sigma_n}{\sigma^{*}}\right)^2 + \left(\frac{\sigma_s}{\tau^{*}}\right)^2 \leq 1
\end{equation}
\noindent where $\sigma_n$, $\sigma_s$ are the actual normal and shear stresses, respectively. $\sigma^{*}$, $\tau^{*}$ are the pure normal and pure shear stresses required for separation, respectively, and are determined experimentally. From Fig.~\ref{fig:bending_fbd}b, we see that $\sigma_n = \frac{T\sin \alpha}{A}$ and $\sigma_s = \frac{T\cos \alpha}{A}$, where $A$ is the area of fastener that experiences stresses. From Salama et al.~\cite{Salama2002},  $A = 8wt$, where $w$ is the fastener width and $t$ is the fastener thickness. 

This yields a two-step process for computing the minimum pressure $P$ for fastener separation. First, compute the maximum tension $T$ within the strength criterion using Eq.~\ref{eq:strength_criterion}. Second, use Eq.~\ref{eq:moment_balance}-\ref{eq:pressure} to solve for the $P$ that results in the maximum $T$.

\section{Experimental Characterization}
In this section, we perform multiple experiments to quantify the shape locking performance of our design. Specifically, we consider the system's ability to hold bends in spite of pressurization of the main body, resist body deflections due to external forces, and form new bends while maintaining previously locked bends. 

\begin{figure}[b]
    \centering
    \vspace{-3mm}
    \includegraphics[width=\linewidth]{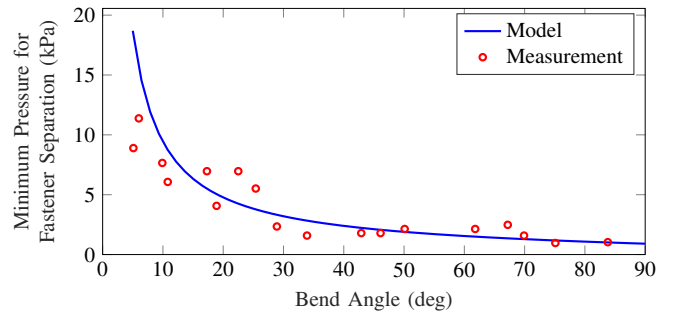}
    \vspace{-6mm}
    \caption{Minimum pressure to cause hook-and-loop fastener separation. Experimental values are in red, and the modeled relationship is in blue.}
    \label{fig:pressure_v_angle}
    \vspace{0mm}
\end{figure}

\subsection{Verifying Bend-Holding Capability}

We experimentally verify our model presented in Sec.~\ref{subec:bend_holding} by measuring the minimum pressure required to initiate separation of hooks and loops on a bend held by the hook-and-loop fastener (Fig.~\ref{fig:pressure_v_angle}). For our experiment, we used an inflated beam with radius 4.0~cm and length 41~cm. The beam also had a strip of loops for locking bends. For a single experiment trial, we first created a bend and locked the bend by applying the hooks onto the loops. We then measured the bend angle. Finally, we increased the pressure until the  hooks began to separate from the beam. We repeated this process for 18 different bend angles. Fig.~\ref{fig:pressure_v_angle} shows the agreement between the measurements and model, and the root mean square error (RMSE) is 2.7~kPa. The RMSE is small relative to the typical body pressure of our vines (about 7~kPa).

\subsection{Beam Stiffness of Unlocked vs. Locked Inflated Beams}
\begin{figure}[t]
    \centering
    \includegraphics[width=2.1in, trim={0 0 0 1.7cm},clip]{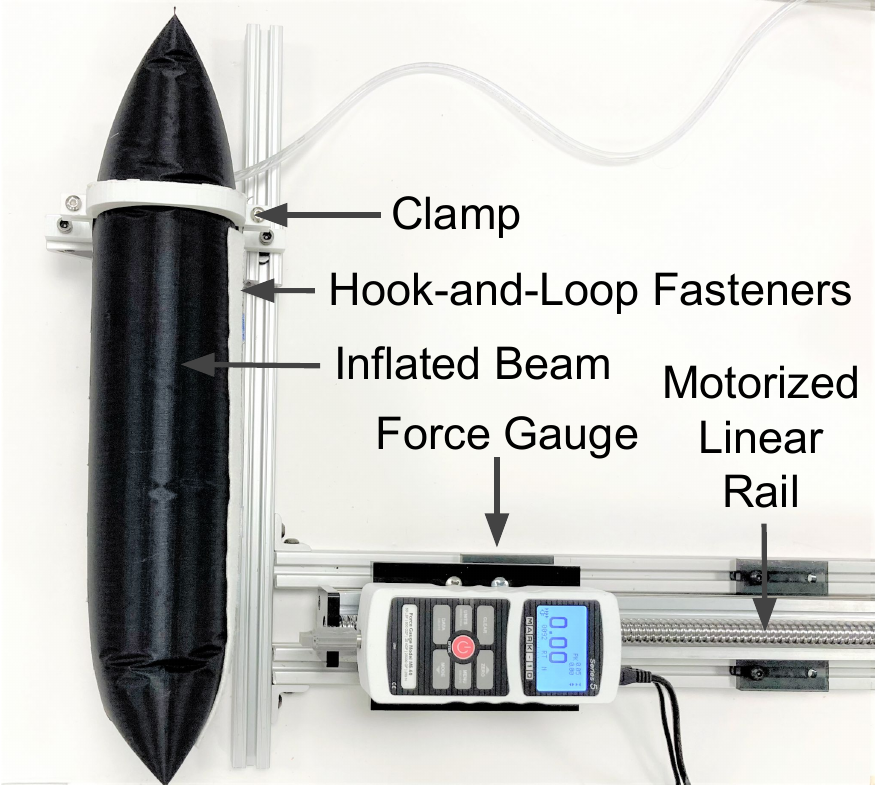}
    \vspace{-2mm}
    \caption{Experimental setup to measure the beam stiffness of locked and unlocked inflated beams.}
    \label{fig:stiffness_setup}
    \vspace{-3mm}
\end{figure}

Here, we characterize the added stiffness that arises from applying hook-and-loop fasteners onto inflated beams. Fig.~\ref{fig:stiffness_setup} shows the experimental setup. We used an inflated beam made out of TPU coated ripstop nylon with length 40~cm, diameter 10.8~cm, and gauge pressure 6.9~kPa. To measure the stiffness, one end of the beam was fixed by a clamp pressed onto an internal ring, and a load on the opposite end of the beam was applied with a force gauge (Mark-10, NY) that moved at constant speed. Stiffness was then computed from this force-displacement data. We executed this process for a beam without fasteners and a beam with a single strip of hook-and-loop fasteners; we refer to these cases as unlocked and locked, respectively. We computed a stiffness of 152~N/m and 199~N/m for the unlocked and locked beams, respectively. The additional 47~N/m in stiffness aids the locked beam in resisting forces from the environment and bending cables. We are interested in increasing this stiffness change in future work.

\subsection{Effect of Cable Tension on Unlocked vs. Locked Bends}
\begin{figure}[t]
    \centering
    \includegraphics[width=\linewidth]{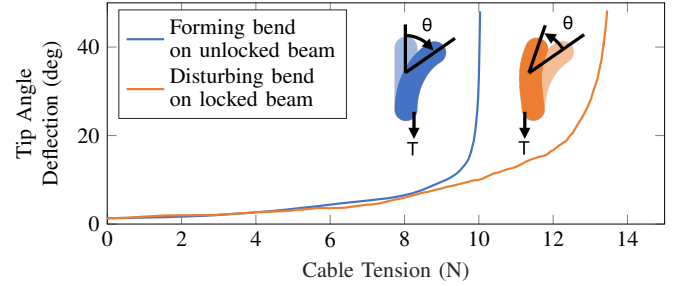}
    \vspace{-6mm}
    \caption{Change in bend angle versus applied cable tension for unlocked and locked beams. The inset illustration shows the two loading cases. For an unlocked beam, applying cable tension T causes an initially straight beam (light blue) to form a bend (dark blue) with a change in tip angle $\theta$. For a locked beam (light orange), applying cable tension T disturbs a locked bend with a change in tip angle $\theta$ (dark orange). The measured data shows that locked bends require more cable tension to cause beam deflection.}
    \vspace{-4mm}
    \label{fig:tension_v_angle}
\end{figure}

Typically, applying tension to cables along a vine robot utilizing a cable and stopper implementation results in a constant curvature along the entire length of the vine. However, by having unlocked and locked regions of a vine, we change how the cables influence the robot configuration. Here we quantify the effect of cable tension on bends by considering the two scenarios that appear during deployment of our system. First, we consider applying cable tension to an unlocked beam. On our full system, the most distal region of the vine is unlocked, and we want applied tension to cause this region to bend so that we steer the direction of vine growth. Second, we consider applying cable tension that opposes an existing, locked bend. On our full system, most of the everted vine is locked, and we do not want subsequently applied tension to cause these regions to change shape. The inset illustrations in Fig.~\ref{fig:tension_v_angle} show these two cases, and in both, a cable tension $T$ causes a change in tip angle $\theta$. The left, blue beam depicts cable tension forming a new bend, and the right, orange beam depicts cable tension disturbing a previously locked bend.

For the two scenarios previously described, we measured the relationship between cable tension and tip angle deflection. Both measurements were taken with an inflated beam with cables and stoppers for bending as well as loops for the locked bend. The cable tension was measured with a force gauge on a linear rail, and the tip angle was measured with a motion capture system (OptiTrack). The measured data is shown in Fig.~\ref{fig:tension_v_angle}. Our measurements verify the general trend that locked bends require more cable tension to cause beam deflection. The exact values for required cable tension would depend on vine robot parameters such as the length of the unlocked region. One limitation in our method is that it does not provide idealized ``locking". For this specific experiment, forming bends requires up to 10~N, which results in a disturbance of 10$\degree$ or less for locked bends. In future work, we plan to further reduce this distrubance.

\section{Demonstrations}
\begin{figure}[b]
    \centering
    \vspace{-3mm}
    \includegraphics[width=\linewidth]{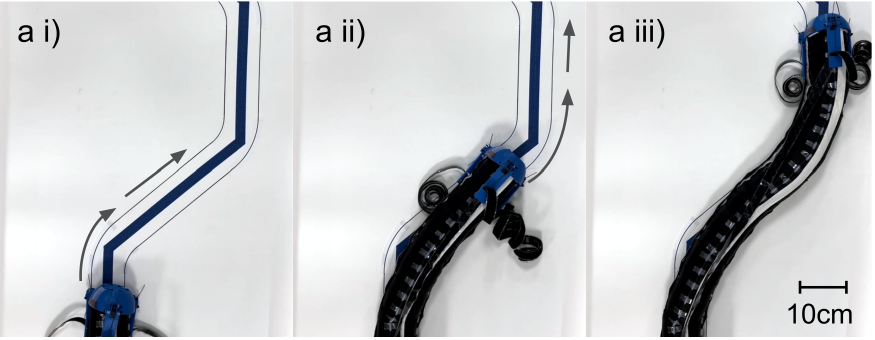}
    \par\vspace{0.5mm}
    \includegraphics[width=\linewidth]{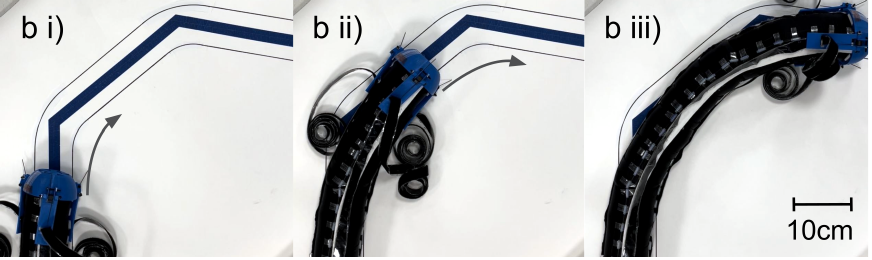}
    \vspace{-6mm}
    \caption{Two demonstrations of our prototype deploying to a desired configuration. Excess hooks are coiled and pulled forward by the tip mount.}
    \vspace{-0mm}
    \label{fig:demo1}
\end{figure}
We used our fabricated prototype to demonstrate the ability of our proposed system to achieve multi-bending via passive shape locking. To show this, we used the same prototype to achieve two different multi-bend configurations. Fig.~\ref{fig:demo1}a shows the first demonstration, in which the vine i) grows to the right and locks that bend in place, ii) grows straight briefly before bending left, and iii) follows the path straight again. This shows the vine's ability to make bends in different directions. In the second demonstration, a different configuration was achieved where the vine made two bends in the same direction. In Fig.~\ref{fig:demo1}b, the vine i) grows straight and bends right, ii) grows straight again, and iii) turns right again. The curvature of the bends in the desired configuration were set based on the kinematics presented in Sec.~\ref{sec:kinematics}. In practice, we found it difficult to achieve the theoretical maximum curvature due to limitations in manual operation.

\begin{figure}[t]
    \centering
    \vspace{2mm}
    \includegraphics[width=\linewidth]{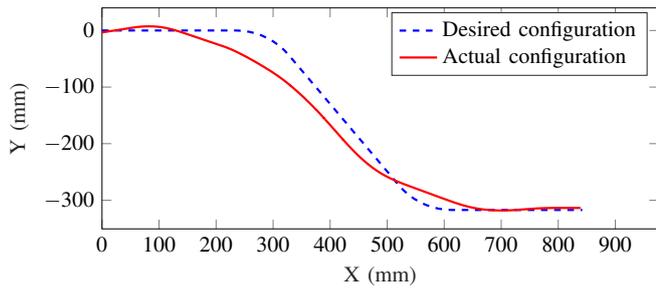}
    \vspace{-7mm}
    \caption{Comparison of the final configuration for our physical proof of concept prototype versus the desired configuration. The prototype was grown and steered by a human operator in real time. The average distance between evenly spaced points on the desired configuration and the corresponding points on the actual configuration is 18~mm. }
    \vspace{-4mm}
    \label{fig:2D_error}
\end{figure}

To measure the ability of our system to achieve a desired configuration, we analyzed the final configuration for the demonstration in Fig.~\ref{fig:demo1}a. We used image processing to extract points along the deployed vine and compared this to the desired configuration Fig.~\ref{fig:2D_error}. To quantify the accuracy, we took evenly spaced points along the deployed and desired configurations and evaluated the Euclidean distance between pairs of points. The average distance (18~mm) is small relative to the diameter of our vine (108~mm).

\section{Conclusion and Future Work}
In this work, we presented a passive shape locking system to enable multi-bend vine robots that are configurable on demand without relying on environment contact and are easily reset and manufactured. We described models that aid in choosing design parameters, presented experiments that quantify the performance of our design, and provided demonstrations that validate the vine robot's ability to achieve accurate multi-bending. In the future, we plan to integrate a retracting mechanism that autonomously removes hooks, explore containment or routing options for the unattached section of hooks, and characterize how well this design scales to longer vines. Our system showed consistent behavior in testing, but we would like to formally characterize robustness and repeatability. Finally, we are interested in demonstrating 3D shapes, exploring other ways to lock bends passively, and investigating how to further rigidize the deployed robot. Our work is a step towards lightweight, low-cost, and low-power reconfigurable deployed inflated structures.

\section*{Acknowledgement}
\noindent The authors thank Alexander K\"{u}bler for design discussions.



\bibliographystyle{IEEEtran}
\bibliography{CHARMBib, paper}

\end{document}